%% file: main.tex
\documentclass[12pt]{iopart}

\usepackage[T1]{fontenc}
\usepackage[utf8]{inputenc}
\usepackage{lmodern}
\usepackage{fullpage}
\usepackage{graphicx}
\usepackage{iopams}
\usepackage{bm}
\usepackage{nicefrac}
\usepackage{hyperref}
\usepackage{xcolor}
\usepackage[export]{adjustbox}

\expandafter\let\csname equation*\endcsname\relax
\expandafter\let\csname endequation*\endcsname\relax
\usepackage{amsmath,amssymb}

\DeclareMathOperator*{\argmin}{arg\,min\;}
\DeclareMathOperator{\prox}{prox}
\DeclareMathOperator{\Var}{Var}

\graphicspath{{figures/}}

\begin{document}
\title{Approximate message-passing for convex optimization with non-separable penalties}
\author{Andre~Manoel$^{1,2}$, Florent~Krzakala$^3$, Bertrand~Thirion$^1$, Gaël~Varoquaux$^1$ and Lenka~Zdeborová$^2$}
\address{$^1$Parietal Team, Inria, CEA, Université Paris-Saclay\\
$^2$Institut de Physique Théorique, CEA, CNRS, Université Paris-Saclay, CNRS\\
$^3$Laboratoire de Physique Statistique, École Normale Supérieure, PSL
University \& Sorbonnes Université, Paris}

\begin{abstract}
    We introduce an iterative optimization scheme for convex objectives
    consisting of a linear loss and a non-separable penalty, based on the
    expectation-consistent approximation and the vector approximate
    message-passing (VAMP) algorithm. Specifically, the penalties we approach
    are convex on a linear transformation of the variable to be determined,
    a notable example being total variation (TV). We describe the
    connection between message-passing algorithms -- typically used for
    approximate inference -- and proximal methods for
    optimization, and show that our scheme is, as VAMP, similar in nature to
    the Peaceman-Rachford splitting, with the important difference that
    stepsizes are set adaptively. Finally, we benchmark the performance of
    our VAMP-like iteration in problems where TV penalties are useful,
    namely classification in task fMRI and reconstruction in tomography, and
    show faster convergence than that of
    state-of-the-art approaches such as FISTA and ADMM in most settings.
\end{abstract}

\input{intro}
\input{methods}
\input{experiments}
\input{conclusion}
\input{ackno}

\appendix
\input{appendices}

\section*{References}

\bibliographystyle{iopart-num}
\bibliography{main}

\end{document}

%% file: intro.tex
\section{Introduction}

Proximal algorithms such as FISTA \cite{beck2009fast} and ADMM
\cite{boyd2011distributed,douglas1956numerical} form these days a
large part of the standard convex optimization toolbox, mostly due to
the fact that they provide different sorts of theoretical guarantees,
such as the convergence to a global minimum of the problem.
However, their use is not so interesting in
some cases. In particular, for non-separable penalties such as total
variation (TV), FISTA requires an inner loop for evaluating the proximal
operator, which takes longer and longer to run as the outer loop iterations
go by.  While variable-splitting algorithms such as ADMM provide a way around
this, they are typically very sensitive to the choice of stepsize, and
determining an optimal stepsize remains an open problem.

An alternative comes from approximate message-passing (AMP) algorithms
\cite{donoho2009message,rangan2011generalized,krzakala2012statistical},
extensions to the belief propagation algorithm
\cite{pearl1988probabilistic} that are capable of dealing with
high-dimensional inference problems. They have shown remarkable success in
applications to compressed sensing, denoising and sparse regression
\cite{vila2013expectation,ziniel2014binary,guo2015near,metzler2016denoising},
but despite that remain disfavored compared to classic optimization
approaches, most prominently due to convergence issues
\cite{rangan2014convergence,caltagirone2014convergence}. The recently
proposed \emph{vector approximate message-passing} (VAMP)
algorithm \cite{rangan2017vector} offers a promising, more robust alternative
to AMP. For a given class of statistical models, notably generalized linear
models (GLMs), different formulations of the VAMP algorithm are capable of
approximating both the maximum a posteriori (MAP) and minimum mean-squared
error (MMSE) estimates. Moreover, these estimates are conjectured to be
asymptotically Bayes-optimal for certain random matrices, and proven to be
in specific cases
\cite{reeves_replica-symmetric_2016,barbier_phase_2017,barbier_mutual_2018}.

Here, we are interested in the evaluation of the MAP estimate, which
is also the solution to a given optimization problem. In particular,
we would like to determine \emph{whether VAMP-like algorithms can provide a
fast alternative to classic optimization approaches} in solving a given class of
convex optimization problems.
VAMP has been typically considered for the minimization of functions consisting
of a quadratic loss and a separable penalty, a setting in which coordinate
descent algorithms are usually much faster than the alternatives, as discussed
in Section \ref{sec:background}.
We thus look at the more interesting (and challenging) case of penalties that
are non-separable---more specifically, separable on disjoint subsets of a
linear transformation of the variable being optimized, a notable example
being total variation (TV) \cite{beck2009fast2,michel2011total}.
For such penalties, proximal algorithms \cite{parikh2014proximal} such as
FISTA and ADMM are considered state-of-the-art.

While VAMP has already been combined with non-separable
denoisers \cite{fletcher2018plug,schniter2017denoising}, there have been no
attempts so far to modify the algorithm to better fit specific types of
non-separable penalties. Moreover, evidence that VAMP has a good performance
in real-world applications is still limited.
The current work addresses these two issues.
In Section \ref{sec:nonsep}, we rederive VAMP for a specific class of
non-separable penalties, by considering the expectation-consistent
(EC) approximation \cite{opper2005expectation} not on the distribution
of the variable of interest, but on that of a linear transform.
Moreover, in Section \ref{sec:experiments}, experiments with total variation
penalties are performed on benchmark datasets, revealing that the adaptation we
propose is competitive with state-of-the-art algorithms.

%% file: methods.tex
\section{Motivation and background}
\label{sec:background}

\subsection{Approximate message-passing}

In what follows, we consider a typical regression problem and denote by
$\bm{y} \in \chi^n$ the response vector (which can represent e.g. labels,
$\chi = \{0, 1\}$, or continuous value $\chi = \mathbb{R}$) and by
$\bm{A} \in \mathbb{R}^{n \times p}$ the feature matrix. 
Then, given a
regularization parameter $\lambda \in \mathbb{R}$ and a convex function $f$,
we look at solving
\begin{equation}
    \hat{\bm{x}} = \argmin_{\bm{x}} \frac12 \| \bm{y} - \bm{A} \bm{x} \|_2^2 +
        \lambda \sum_j f (x_j),
    \label{eq:prob_sep}
\end{equation}
The approximate message-passing (AMP) algorithm in
\cite{donoho2009message} aims at solving (\ref{eq:prob_sep}) by means of the
following iteration \begin{align}
    &\bm{x}^{t + 1} = \eta_{\lambda \sigma^t} (\bm{x}^t + \bm{A}^T \bm{z}^t),
        \label{eq:amp1} \\
    &\bm{z}^t = \bm{y} - \bm{A} \bm{x}^t + \frac{1}{\alpha} \bm{z}^{t - 1}
        \big\langle \nabla \eta_{\lambda \sigma^t} (\bm{x}^t + \bm{A}^T \bm{z}^t) \big\rangle
        \label{eq:amp2},
\end{align}
where $\alpha = n / p$, $\sigma^t$ is an empirical estimate on the variance of
$\bm{x}$, $\langle \bm{a} \rangle = \frac{1}{n_{\bm{a}}} \sum_{i = 1}^{n_{\bm{a}}}
a_i$ is the empirical average over the elements of $\bm{a} \in
\mathbb{R}^{n_{\bm{a}}}$, and
\begin{equation}
    \eta_{\lambda \sigma^t} (v) = \prox_{\lambda \sigma^t f} (v) =
        \argmin_{x} \left\{ \lambda \sigma^t f(x) + \frac{1}{2} (x - v)^2
        \right\},
\end{equation}
is a function applied elementwise. In practice, the variance estimate is
updated using \cite{donoho2010message} $\sigma^{t + 1} = 1 + \sigma^{t}
\langle \nabla \eta_{\lambda \sigma^t} \rangle / \alpha$.

AMP's form is close to that of a proximal gradient descent
\cite{parikh2014proximal}, being in fact very similar to the iterative
soft thresholding algorithm (ISTA), with, however, an additional term.
The extra term in (\ref{eq:amp2}), known as Onsager reaction term,
comes from a second order correction derived using statistical physics
techniques \cite{thouless1977solution,mezard1989space}, and has been
shown to improve the iteration in synthetic data experiments. An instance
of this can be seen in the example of subsection \ref{sec:example}.

An important assumption is made in deriving AMP: that the entries of $\bm{A}$
are i.i.d., have zero mean, and variance that scales as $1 / n$. The
iteration typically fails to converge whenever this assumption is not met
\cite{rangan2014convergence,caltagirone2014convergence}.
A series of works have tried to improve the convergence properties of AMP by
means of heuristic modifications \cite{vila2015adaptive,manoel2015swept},
typically resulting in slower algorithms that provide few theoretical
guarantees.

A promising alternative is the vector AMP (VAMP) algorithm, recently
introduced by \cite{rangan2017vector,ma2017orthogonal} as a more robust message-passing
scheme. It is shown to converge, in the large $n$ limit, for a wider class
of matrices $\bm{A}$ -- namely that of rotationally-invariant random matrices. 
As AMP, VAMP provides an iteration which upon convergence
returns a solution to (\ref{eq:prob_sep}); it uses, however, a
different framework than AMP, known as expectation-consistent (EC)
approximation \cite{opper2005expectation}, which is briefly described in
\ref{sec:ec}. For problem (\ref{eq:prob_sep}), the VAMP iteration
reads
\begin{align}
    &\bm{x}^{t} = (\bm{A}^T \bm{A} + \rho^t I_n)^{-1} (\bm{A}^T \bm{y} + \bm{u}^{t}), \qquad
    &&\bm{z}^{t} = \eta_{\lambda \sigma_x^t / (1 - \sigma_x^t \rho^t)} \bigg(
        \frac{\bm{x}^t - \sigma_x^t \bm{u}^t}{1 - \sigma_x^t \rho^t} \bigg),
        \label{eq:vamp1} \\
    &\bm{u}^{t + 1} = \bm{u}^t + (\bm{z}^t / \sigma_z^t - \bm{x}^t / \sigma_x^t), \qquad
    &&\rho^{t + 1} = \rho^t + (1 / \sigma^t_z - 1 / \sigma^t_x). \label{eq:vamp2}
\end{align}
While AMP is close to ISTA, the iteration above resembles that of another
popular optimization algorithm, the alternating directions method of
multipliers (ADMM) \cite{boyd2011distributed}. This connection is discussed
in subsection \ref{sec:varsplit}.

The $\sigma_x^t$ and $\sigma_z^t$ are estimates of the variance of
$\bm{x}$ and of the \emph{split variable} $\bm{z}$, and are obtained from
\begin{equation}
    \sigma^t_x = \frac{1}{n} \Tr (\bm{A}^T \bm{A} + \rho^t I_n)^{-1}, \kern4em
    \sigma^t_z = \frac{\sigma_x^t}{1 - \sigma_x^t \rho^t} \, \bigg\langle
        \nabla \eta_{\lambda \sigma_x^t / (1 - \sigma_x^t \rho^t)} \bigg( \frac{\bm{x}^t -
        \sigma_x^t \bm{u}^t}{1 - \sigma_x^t \rho^t} \bigg) \bigg\rangle.
    \label{eq:vamp3}
\end{equation}
Iterations (\ref{eq:vamp1}) and (\ref{eq:vamp3}) can be seen as evaluating
the means and variances of two distinct distributions, commonly referred to
as \emph{tilted} distributions in the context of EC. By forcing these
distributions to concentrate around their modes, the means become proximal
operators, whereas the variances come from the curvature of
Moreau envelopes of the loss and the penalty. This is detailed in
\ref{sec:ec}.

The matrix inversion in (\ref{eq:vamp1}) and (\ref{eq:vamp3}) can be quite costly;
however, the iteration can be modified to use the SVD decomposition 
of $\bm{A} = U \Sigma V$ \cite{rangan2017vector}, in which case $\sigma^t_x =
\frac{1}{n} \sum_{i = 1}^n \frac{1}{\Sigma^2_{ii} + \rho^t}$.
In particular, if $n \ll p$, the Woodbury formula can be combined with the
SVD to provide an efficient iteration, which is used in our experiments (see
\ref{sec:app_nllp} for more details).

Unless otherwise mentioned, we shall always assume that $\bm{x}^0$, $\bm{z}^0$
and $\bm{u}^0$ are initialized to all-zero vectors, while $\rho^0 = 1$
and $\sigma_x^0 = \sigma_z^0 = 1/2$.

\subsubsection{Example: $\ell_1$ regularization on synthetic data}
\label{sec:example}

\begin{figure}[t]
    \centering
    \includegraphics[width=\textwidth]{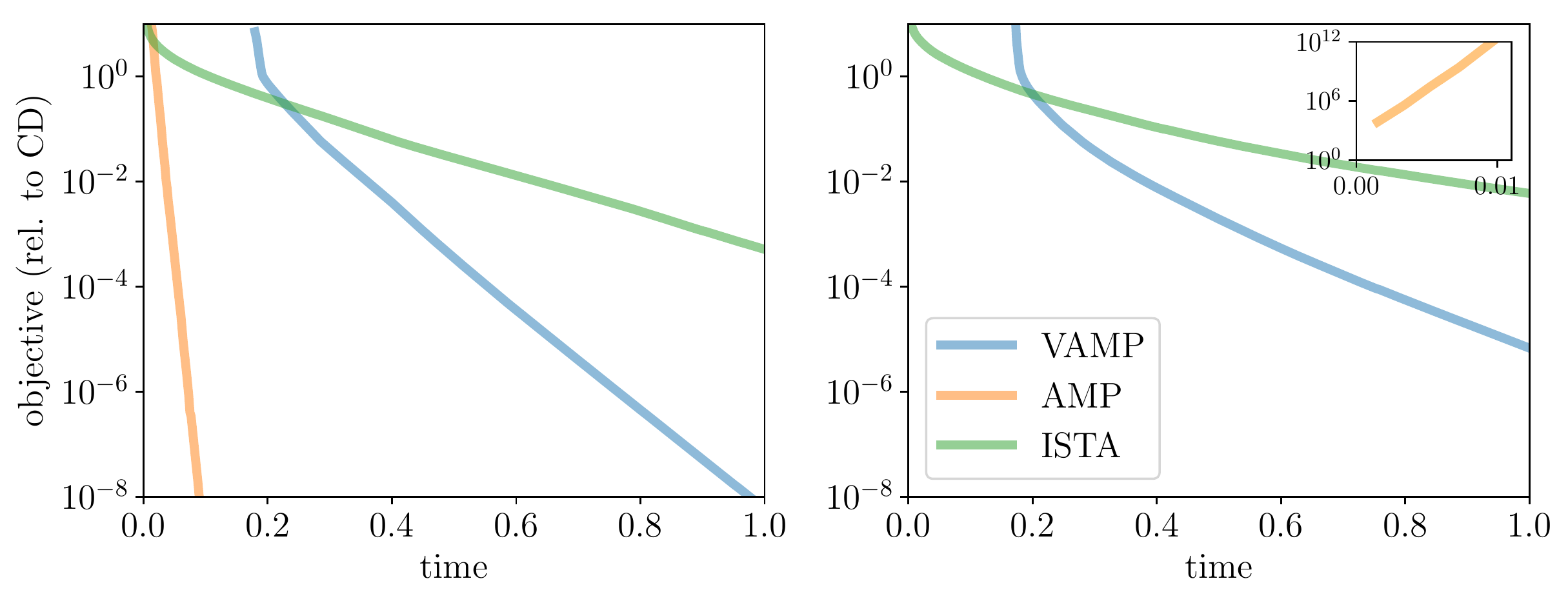}
    \caption{Comparison between AMP and VAMP for sparse regression on synthetic data,
        with $n = 600$, $p = 2000$ and $\rho = 0.1$, $\sigma = 10^{-5}$,
        $\lambda = 1$. Objectives are relative to that of scikit-learn's
        coordinate descent \cite{scikit-learn}, which runs in a much shorter
        time (around $5 \cdot 10^{-4}$s). Left: for
        Gaussian i.i.d. matrices, VAMP performs worse than AMP, both in terms
        of convergence rate and preprocessing time. Right: for a matrix
        which is not Gaussian i.i.d., but instead the product of two
        Gaussian i.i.d. matrices $\bm{A} = \bm{U} \bm{V}^T$, AMP quickly
        diverges, as shown in the inset, while VAMP keeps providing a good
        performance. In this example, both $\bm{U}$ and $\bm{V}$ have $r =
        600$ columns.}
    \label{fig:amp_vs_vamp}
\end{figure}

In order to compare the AMP iteration (\ref{eq:amp1})-(\ref{eq:amp2}) to the
VAMP one, (\ref{eq:vamp1})-(\ref{eq:vamp2}), we perform a simple experiment on
synthetic sparse data. We first sample $\bm{x}$ from a Bernoulli-Gaussian
distribution, $P_X (\bm{x}) = \prod_j [\rho \, \mathcal{N} (x_j; 0, 1) + (1 -
\rho) \, \delta (x_j)]$, then generate a matrix $\bm{A}$ by doing $\bm{A}_{ij} \sim
\mathcal{N} (\bm{A}_{ij}; 0, 1/n)$, to finally obtain measurements as $\bm{y} \sim
\mathcal{N} (\bm{y}; \bm{A} \bm{x}, \sigma^2 I_n)$.

We try to recover $\bm{x}$ by using a $\ell_1$ penalty, $f (x_j) = |x_j|$.
In that case\footnote{We denote $(\cdot)_+ = \max (\cdot, 0)$.},
$\eta_{\lambda \sigma^t} (x) = \Big(1 - \frac{\lambda \sigma^t}{|x|}\Big)_+
x$, also known as the soft thresholding operator. In Figure
\ref{fig:amp_vs_vamp} (left), we study how fast each of the schemes performs this
task.
We look at three different schemes: iterative soft thresholding (ISTA)
\cite{donoho1995noising}, the original AMP iteration
(\ref{eq:amp1})-(\ref{eq:amp2}), and the VAMP iteration
(\ref{eq:vamp1})-(\ref{eq:vamp3}). Compared to AMP and ISTA, VAMP
has a larger preprocessing time, since it relies on the eigendecomposition
of $\bm{A} \bm{A}^T$, as detailed in \ref{sec:app_nllp}. Moreover,
in this artificial setting, its convergence rate is smaller than that of
AMP.

Yet, such Gaussian matrices are precisely those assumed in the
derivation of AMP. If a different matrix is used, say $\bm{A} = \bm{U}
\bm{V}^T$ with both $\bm{U}$ and $\bm{V}$ Gaussian, then the AMP
iteration quickly diverges, whereas VAMP continues to work (see Figure
\ref{fig:amp_vs_vamp}, right).
One verifies in practice that VAMP converges for a broad class of
matrices, and is actually proven to converge, in the large $n$ limit, for
right orthogonally-invariant random matrices \cite{rangan2017vector}. This
motivates us to perform more detailed experiments with VAMP, as well as to
adapt it to our needs.

Also notably, all schemes used in Figure \ref{fig:amp_vs_vamp} are much slower
than coordinate descent, which is currently considered the state-of-the-art for optimization
with separable penalties. For this reason, we shall consider the more
interesting case of non-separable penalties in Section \ref{sec:nonsep}.

\subsection{Variable splitting methods}
\label{sec:varsplit}

\begin{figure}[t]
    \centering
    \includegraphics[width=\textwidth]{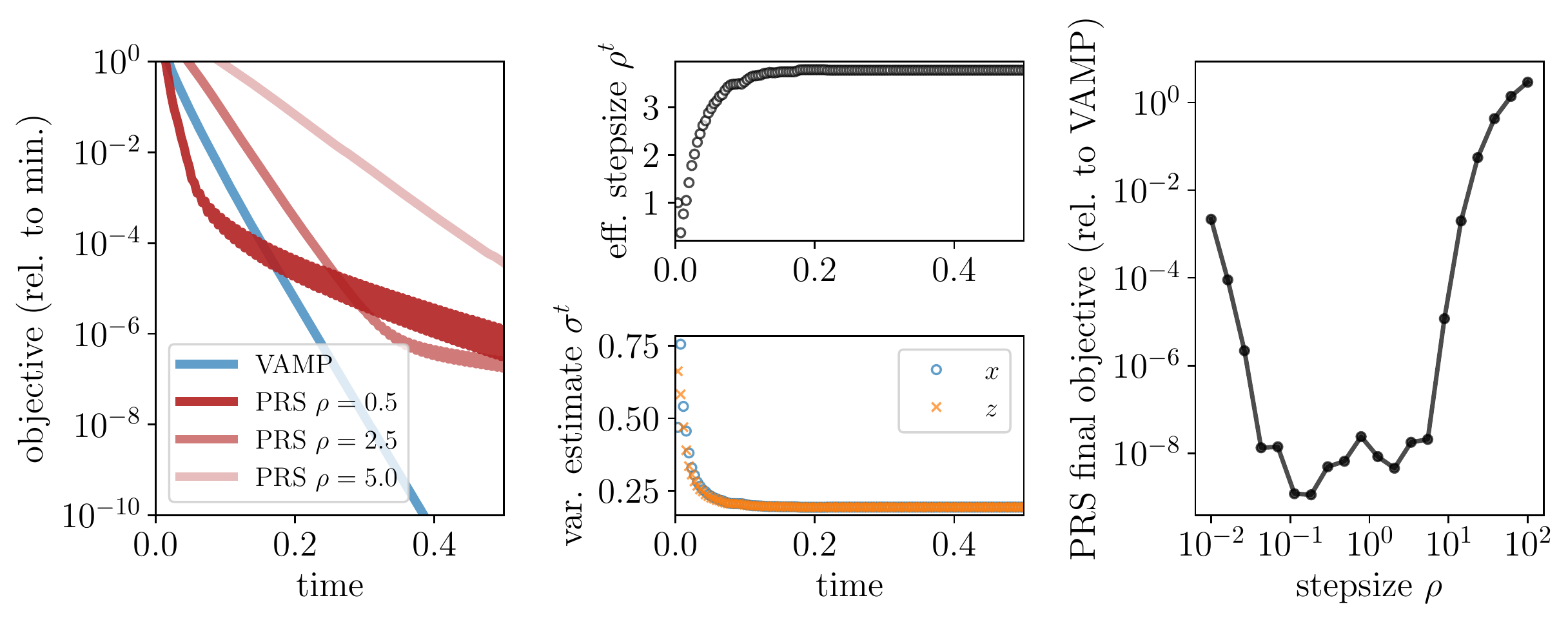}
    \caption{Comparison between VAMP and PRS algorithms for sparse regression on
        synthetic data, with $n = 1200$, $p = 2000$ and $\rho = 0.2$,
        $\sigma^2 = 10^{-5}$, $\lambda = 1$.  Left: PRS behavior is highly
        dependent on the value chosen for the stepsize $\rho$, whereas for
        VAMP no parameter needs to be set.  Center: effective stepsize (top)
        and variance estimates (bottom) provided by VAMP.  Right: difference
        between PRS and VAMP objectives after 250 iterations.}
    \label{fig:vamp_vs_pr}
\end{figure}

We now turn to a class of optimization algorithms known as \emph{variable
splitting methods}. The main very basic idea is to minimize a sum of two functions such as
(\ref{eq:prob_sep}) by writing each of them as a function of a different
variable, and then enforcing these variables to assume the same value. This is
done by means of the following augmented Lagrangian
\begin{equation}
    \mathcal{L} (\bm{x}, \bm{z}, \bm{u}) = \frac12 \| \bm{y} - \bm{A} \bm{x} \|_2^2 +
        \lambda \sum_j f(z_j) \underbrace{- \bm{u}^T (\bm{x} - \bm{z}) +
        \frac{\rho}{2} \| \bm{x} - \bm{z} \|^2_2}_{\frac{\rho}{2} \| \bm{x}
        - \bm{z} - \frac{\bm{u}}{\rho} \|_2^2 - \frac{1}{2 \rho} \| \bm{u}
        \|_2^2},
    \label{eq:lag_sep}
\end{equation}
One should minimize the expression above with respect to $\bm{x}$ and $\bm{z}$,
and maximize its dual $\mathcal{L}^\ast = \min_{\bm{x}, \bm{z}} \mathcal{L}
(\bm{x}, \bm{z}, \bm{u})$ with respect to the Lagrange multiplier $\bm{u}$,
see e.g. \cite{boyd2004convex,nocedal2006numerical}.  The alternating
direction method of multipliers (ADMM)
\cite{boyd2011distributed,douglas1956numerical} accomplishes this by means
of the following iteration
\begin{align}
    &\bm{x}^{t} = \argmin_{\bm{x}} \; \mathcal{L} (\bm{x}, \bm{z}^{t}, \bm{u}^{t}) = 
        \prox_{L / \rho} \, (\bm{z}^{t} + \nicefrac{\bm{u}^{t}}{\rho}), \label{eq:admm1} \\
    &\bm{z}^{t} = \argmin_{\bm{z}} \; \mathcal{L} (\bm{x}^{t}, \bm{z}, \bm{u}^{t}) =
        \prox_{R / \rho} \, (\bm{x}^{t} - \nicefrac{\bm{u}^{t}}{\rho}), \label{eq:admm2} \\
    &\bm{u}^{t + 1} = \bm{u}^{t} + \rho (\bm{z}^{t} - \bm{x}^{t}). \label{eq:admm3}
\end{align}
where we have defined $L (\bm{x}) = \frac12 \| \bm{y} - \bm{A} \bm{x} \|_2^2$ and $R
(\bm{z}) = \lambda \sum_j f(z_j)$. Equations (\ref{eq:admm1}) and
(\ref{eq:admm2}) perform two sequential gradient descent steps in
$\mathcal{L}$, first for $\bm{x}$ then $\bm{z}$, whereas equation
(\ref{eq:admm3}) is effectively a gradient ascent step in $\mathcal{L}^\ast
(\bm{u})$, with stepsize set so as to satisfy the optimality conditions
$\nabla_{\bm{x}, \bm{z}} \mathcal{L} = 0$.

The strategy employed by ADMM is empirically known as \emph{Douglas-Rachford} splitting.
Different strategies, with the same fixed points but slightly different
updates, can be employed. For instance, the \emph{Peaceman-Rachford}
splitting (PRS) consists in the following
equations \cite{gabay1983applications,peaceman1955numerical}
\begin{align}
    &\bm{x}^{t} = \prox_{L / \rho} \; (\nicefrac{\bm{u}^{t}}{\rho}), \label{eq:prs1} \\
    &\bm{z}^{t} = \prox_{R / \rho} \; (2 \bm{x}^{t} - \nicefrac{\bm{u}^{t}}{\rho}), \label{eq:prs2} \\
    &\bm{u}^{t + 1} = \bm{u}^{t} + 2 \rho (\bm{z}^{t} - \bm{x}^{t}). \label{eq:prs3}
\end{align}
While PRS is known to often lead to lower objectives faster than
ADMM \cite{gabay1983applications,he2014strictly}, it is provably convergent
under stronger assumptions, and in practice the objective
often does not decrease monotonically.

Writing the proximal operators explicitly in the PRS iteration
(\ref{eq:prs1}) and (\ref{eq:prs2}) yields
\begin{equation}
    \bm{x}^{t} = (\bm{A}^T \bm{A} + \rho I_n)^{-1} (\bm{A}^T \bm{y} + \bm{u}^t), \kern4em
    \bm{z}^{t} = \eta_{\lambda / \rho} (2 \bm{x}^t - \bm{u}^t / \rho).
\end{equation}
Interestingly, setting $\sigma_x^t = \sigma_z^t = \frac{1}{2 \rho}$ in the
VAMP iteration (\ref{eq:vamp1})-(\ref{eq:vamp2}) leads to the PRS iteration
above; VAMP can thus be seen as a PRS variant where the stepsize is set
adaptively. This remark has been first made in \cite{fletcher2016expectation}.

\vspace{3ex}

We thus see that VAMP is more robust than AMP, converging in situations
where AMP does not, as shown in Figure \ref{fig:amp_vs_vamp}. It also has a
significant advantage over variable-splitting methods, since no stepsize
needs to be set.
While neither VAMP nor variable-splitting methods are competitive with the
state-of-the-art in the particular setting of separable penalties,
we now turn to non-sperable penalties, which are more challenging regarding
efficiency and convergence control \cite{dohmatob2014benchmarking}, yet very
important in practical applications.

\section{Non-separable penalties}
\label{sec:nonsep}

Consider a penalty which is separable not on $\bm{x}$ but on disjoint subsets of
a linear transformation $\bm{K} \bm{x}$, with $\bm{K} \in \mathbb{R}^{r \times n}$
\begin{equation}
    \hat{\bm{x}} = \argmin_{\bm{x}} \frac12 \| \bm{y} - \bm{A} \bm{x} \|_2^2 +
        \lambda \sum_{k = 1}^r f\big( (\bm{K} \bm{x})_k \big),
    \label{eq:prob_nonsep}
\end{equation}
for any convex function $f$. Note that $k$ does not necessarily index a
single component of $\bm{K} \bm{x}$, but more generally a subset of components;
moreover, each component belongs to a single subset.
An example is the so-called total variation (TV)
penalty \cite{beck2009fast2,michel2011total}, used to enforce spatial
regularity -- typically flat regions separated by sharp transitions
\begin{equation}
	R_{\rm TV} (\bm{x}) = \sum_{ij} \sqrt{(x_{i, j} - x_{i, j-1})^2
		+ (x_{i, j} - x_{i-1, j})^2},
\end{equation}
where we have indexed the vector $\bm{x}$ by two components $i, j$ such that
$x_k = x_{i, j}$ for $i = \lfloor k / L \rfloor$ and $j = k \mod L$. By
defining a gradient operator $\nabla$
\begin{equation}
	(\nabla \bm{x})_k = \begin{pmatrix} x_{i, j+1} - x_{i,j} \\ x_{i + 1, j} -
		x_{i, j} \end{pmatrix},
\end{equation}
the 2D TV penalty can be conveniently written as $R_{\rm TV} = \sum_{k = 1}^{2n}
\| (\nabla \bm{x})_k \|_2 \equiv \| \nabla \bm{x} \|_{2, 1}$. Different boundary
conditions can be used; in our experiments, we assume periodic boundary conditions.

While we define it above for two dimensions, the TV operator is easily generalizated
to lattices in three or more dimensions. Notice moreover that we are dealing
with the \emph{isotropic} or rotation-invariant TV operator, instead of the
anisotropic one $\| \nabla \bm{x} \|_{1, 1}$.

Variable splitting algorithms can be easily adapted to that case, by
using a split variable $\bm{z} = \bm{K} \bm{x}$; the augmented Lagrangian
(\ref{eq:lag_sep}) becomes
\begin{equation}
    \mathcal{L} (\bm{x}, \bm{z}, \bm{u}) = \frac12 \| \bm{y} - \bm{A} \bm{x} \|_2^2 +
        \lambda \sum_{k = 1}^r f(z_k) - \bm{u}^T (\bm{K} \bm{x} - \bm{z}) +
        \frac{\rho}{2} \| \bm{K} \bm{x} - \bm{z} \|^2_2.
\end{equation}
and the PRS iteration is now written
\begin{align}
    &\bm{x}^{t} = (\bm{A}^T \bm{A} + \rho \bm{K}^T \bm{K})^{-1} (\bm{A}^T \bm{y} + \bm{K}^T \bm{u}^t), \label{eq:prs21} \\
    &\bm{z}^{t} = \eta_{\lambda / \rho} (2 \bm{K} \bm{x}^t - \bm{u}^t / \rho), \label{eq:prs22} \\
    &\bm{u}^{t + 1} = \bm{u}^{t} + 2 \rho (\bm{z}^{t} - \bm{K} \bm{x}^{t}). \label{eq:prs23}
\end{align}
In particular for TV, $\bm{K} = \nabla$ and the $\eta$ function is given by the group
soft thresholding operator
\begin{equation}
    \eta_{\lambda / \rho} (\bm{v}) = \bigg(1 - \frac{\lambda / \rho}{\| \bm{v}
    \|_2}\bigg)_+ \, \bm{v}.
\end{equation}

\subsection{VAMP for non-separable penalties}

The following VAMP-like iteration, which we derive in \ref{sec:adapt},
can be used in order to solve problem \ref{eq:prob_nonsep}
\begin{align}
    &\bm{x}^{t} = (\bm{A}^T \bm{A} + \rho^t \bm{K}^T \bm{K})^{-1} (\bm{A}^T \bm{y} + \bm{K}^T \bm{u}^{t}),
    &&\sigma^t_x = \frac{1}{n} \Tr \big[\bm{K} (\bm{A}^T \bm{A} + \rho^t \bm{K}^T \bm{K})^{-1} \bm{K}^T \big],
        \label{eq:tvamp1} \\
    &\bm{z}^{t} = \eta_{\frac{\lambda \sigma_x^t}{1 - \sigma_x^t \rho^t}} \bigg(
        \frac{\bm{K} \bm{x}^t - \sigma_x^t \bm{u}^t}{1 - \sigma_x^t \rho^t} \bigg),
    &&\sigma^t_z = \frac{\sigma_x^t}{1 - \sigma_x^t \rho^t} \, \bigg\langle
        \nabla \eta_{\frac{\lambda \sigma_x^t}{1 - \sigma_x^t \rho^t}} \bigg( \frac{\bm{K} \bm{x}^t -
        \sigma_x^t \bm{u}^t}{1 - \sigma_x^t \rho^t} \bigg) \bigg\rangle,
        \label{eq:tvamp2} \\
    &\bm{u}^{t + 1} = \bm{u}^t + (\bm{z}^t / \sigma_z^t - \bm{K} \bm{x}^t / \sigma_x^t),
    &&\rho^{t + 1} = \rho^t + (1 / \sigma^t_z - 1 / \sigma^t_x).
    \label{eq:tvamp3}
\end{align}
Upon convergence, it is verified that
$\bm{z} = \bm{K} \bm{x}$ and $\sigma_x = \sigma_z$. Once again, if the variances
remain fixed throughout the iteration, $\sigma_x^t = \sigma_z^t = \frac{1}{2 \rho}$,
the PRS iteration (\ref{eq:prs21})-(\ref{eq:prs23}) is recovered.
Moreover, the Woodbury formula can also be applied, leading to a faster iteration
whenever $n \ll p$. The iteration is particularly fast for the TV matrix $\bm{K} = \nabla$,
for which $\bm{K}^T \bm{K} = \Delta$ is a circulant matrix and thus diagonal in the
Fourier basis, see \ref{sec:simp_tv}.

While we consider convex $f$ functions, we never use this assumption
explicitly. In other words, the iteration above could, in principle, also
be used when $f$ is non-convex.

\subsection{Enforcing contractivity}

A well-known issue with PRS is that, differently from ADMM, the iteration
mapping might not be contractive \cite{he2002new,he2014strictly}, which might
lead to lack of convergence. The iteration we propose,
(\ref{eq:tvamp1})-(\ref{eq:tvamp3}), suffers from the same issue. A way
of enforcing contractivity, done for PRS in \cite{he2014strictly}, is by
including a underdetermined relaxation factor $0 < \gamma < 1$ in the update
of the Lagrange multiplier $\bm{u}$, that is
\begin{equation}
    \bm{u}^{t + 1} = \bm{u}^t + \gamma (\bm{z}^t / \sigma_z^t - \bm{K} \bm{x}^t /
        \sigma_x^t).
\end{equation}
For consistency, we also include the same parameter in the update of the
stepsize $\rho$
\begin{equation}
    \rho^{t + 1} = \rho^t + \gamma (1 / \sigma_z^t - 1 / \sigma_x^t).
\end{equation}

While this modification improves the convergence properties of the algorithm,
it also makes us need to specify an extra parameter $\gamma$. This parameter
could in principle be set adaptively, as it has been done for other AMP
algorithms \cite{vila2015adaptive}. In our experiments, however, we leave
this parameter fixed at a small enough value chosen arbitrarily, $\gamma =
0.6$.

%% file: experiments.tex
\section{Numerical experiments}
\label{sec:experiments}

In this section we solve problem (\ref{eq:prob_nonsep}) for a 2D total
variation penalty, $\bm{K} = \nabla$, and different choices of $\{\bm{y}, \bm{A}\}$.
We compare the iteration we propose, (\ref{eq:tvamp1})-(\ref{eq:tvamp3}), to
three other standard algorithms: ADMM with an adaptive
stepsize \cite{xu2017adaptive}; a FISTA \cite{beck2009fast} adaptation known as
FAASTA \cite{varoquaux2015faasta}, as implemented in\footnote{We have performed
a small modification to this implementation, so as to allow periodic boundary
conditions.} the Nilearn package \cite{abraham2014machine}; and the
Peaceman-Rachford splitting, recovered by leaving variances fixed in our own
implementation.

We work in the high-dimensional regime with $n \ll p$ and thus use the numerical tricks described
in \ref{sec:app_nllp}, combining the Woodbury formula to the
eigendecompositions of $\bm{A} \bm{A}^T$ and $\bm{K}^T \bm{K}$ to speed up the evaluation of the
proximal step in VAMP, PRS and ADMM. In all experiments, we set the relaxation
factor to $\gamma = 0.6$ in VAMP, and to $\gamma = 0.95$ in PRS.

All experiments have been performed on a machine with a 12-core Intel i7-8700K processor
and 32GB of RAM, using Numpy and a multi-CPU implementation of BLAS (OpenBLAS).

\subsection{Classification in task fMRI}

\begin{figure}[ht]
    \centering
    \includegraphics[width=0.53\textwidth,valign=c]{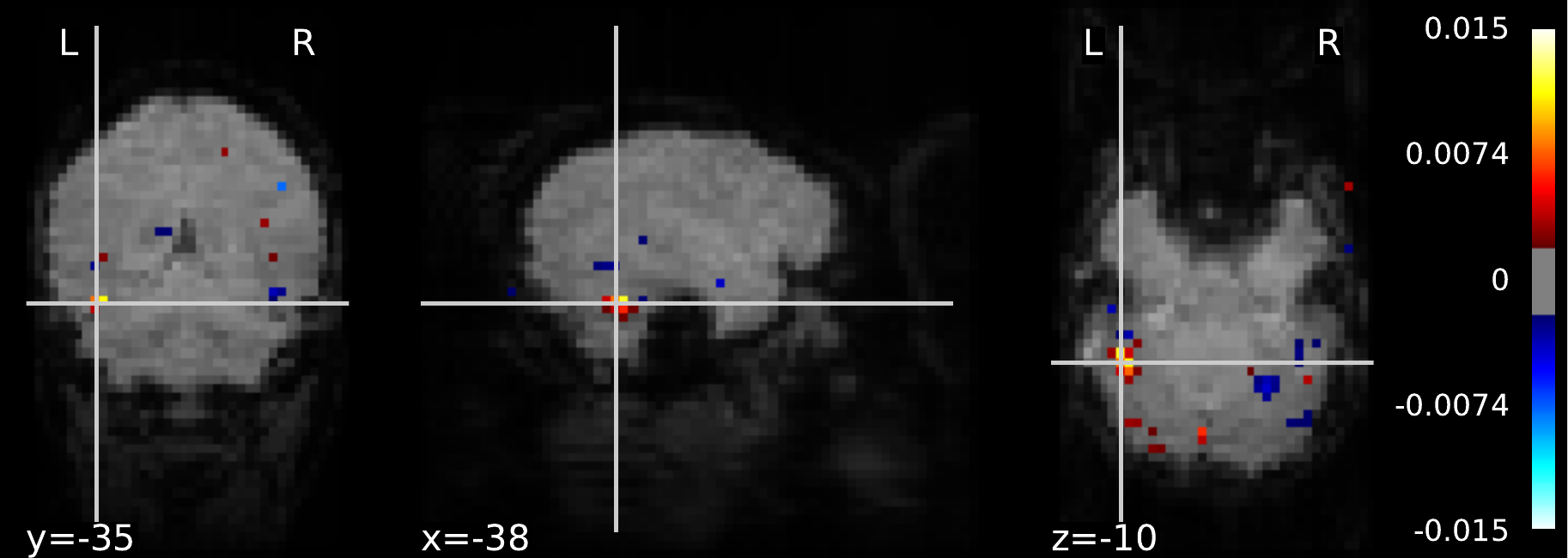} \qquad
    \includegraphics[width=0.4\textwidth,valign=c]{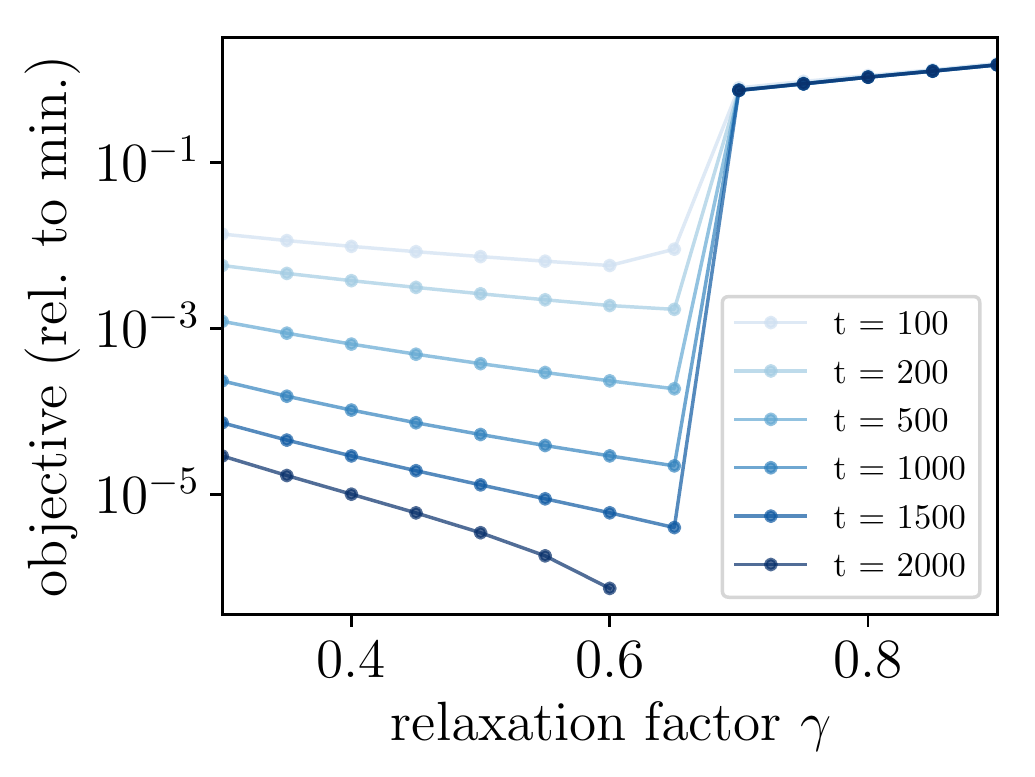}
    \caption{\emph{Left}: weight map obtained by VAMP in classifying
    ``face'' vs. all. Intensities are thresholded at 0.002. \emph{Right}:
    objective in classifying ``face'' vs. all at different numbers of
    iterations, for different choices of relaxation factor. If the
    relaxation factor is set small enough, there is no noticeable difference
    in performance.}
    \label{fig:haxby_tv-weights}
\end{figure}

In our first numerical experiment, we approach a multi-class classification problem,
namely a task fMRI one: each row of $\bm{A} \in \mathbb{R}^{n \times p}$ is given
by an fMRI image, while the entries of $\bm{y} \in \mathbb{R}^n$ tell us what
task was being performed by the subject at the time of acquisition. We use
the Haxby dataset \cite{haxby2001distributed}, which is commonly used for
benchmarks and contains $n = 1452$ samples of $p = 40 \times 64 \times 64 =
163840$ features each. The classes refer to what image the subject was
observing at the time of acquisition. We perform one vs. all classification
for three of these classes: ``face'', ``house'' and ``chair'', each
associated to 108 of the 1452 samples.

While $p = 163840$, many of the voxels are zero-valued and thus a mask can be applied,
leading to a new feature matrix with only $39912$ columns. One can pass this smaller
matrix to FISTA instead of the full one, as it is capable of dealing with
such masks. However, the remaining methods require the full matrix, at least
in the way we have implemented them.

\begin{figure}[p]
    \includegraphics[width=\textwidth]{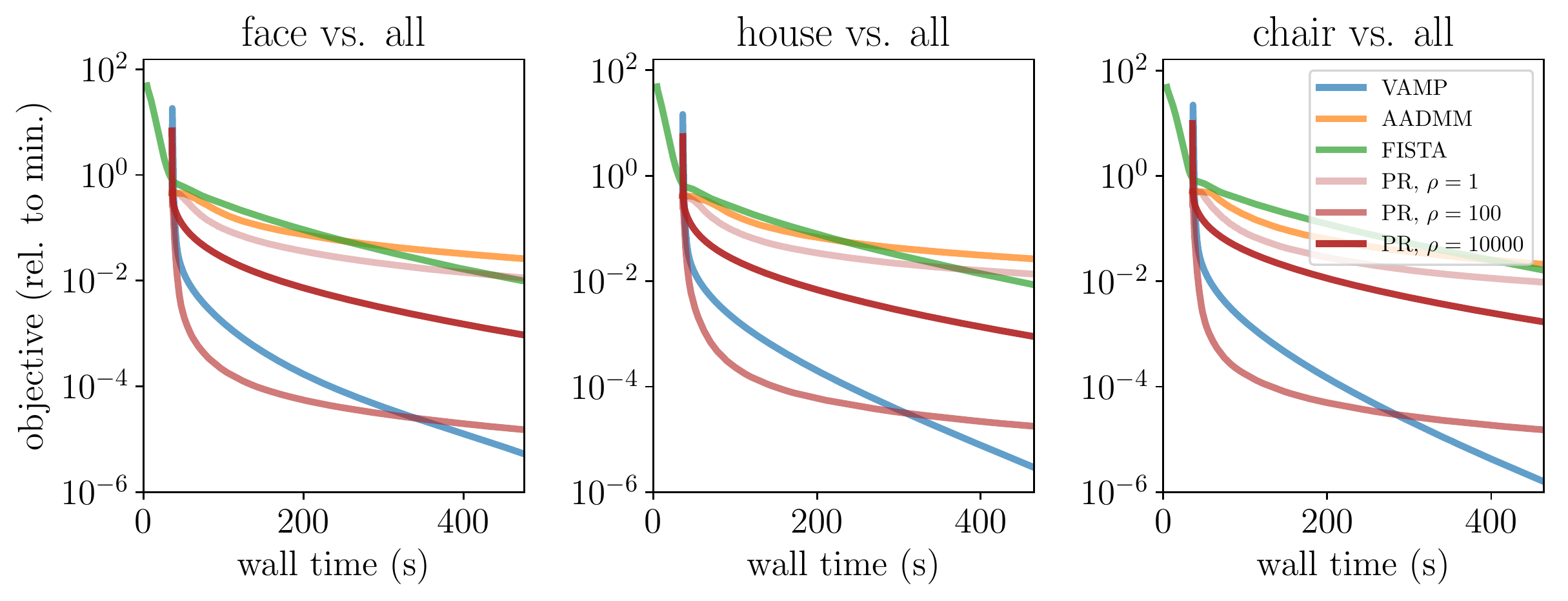}
    \caption{Classification in task fMRI using the Haxby dataset, $n = 1452$
    and $p = 136840$. Since there are multiple classes, we perform one vs.
    all classification for three specific ones: ``face'', ``house'' and
    ``chair''. All schemes seek to solve the same convex optimization problem
    (\ref{eq:prob_nonsep}) with a TV penalty, $\bm{K} = \nabla$, and $\lambda =
    0.1$. For all three classes considered, VAMP achieves a lower objective
    faster than all other schemes. VAMP, AADMM and PR have a late start due
    to preprocessing time, during which the eigendecomposition of $\bm{A}^T \bm{A}$ is
    evaluated.}
    \label{fig:haxby_tv}
\end{figure}

Two of the parameters are fixed: the regularization $\lambda = 0.1$
and the relaxation $\gamma$, which is set to 0.6 for VAMP and to 0.95 for
PRS. In this setting, VAMP achieves smaller values of the objective faster
than all other methods and for all the three classes considered, as shown in
Figure \ref{fig:haxby_tv}. The weight map obtained in ``face'' classification
is displayed in Figure \ref{fig:haxby_tv-weights}.

In Figure \ref{fig:haxby_tv-weights} (right), we study how the performance
of our scheme changes in this experiment as a function of $\gamma$. We can
see the performance does not change considerably as long as $\gamma$ is set
small enough. While this has been often observed by us, further
experiments must be performed to ensure such behavior is indeed recurrent.
If that is the case, it should be straightforward to devise an adaptive
scheme for $\gamma$.


\begin{figure}[p]
    \centering
    \includegraphics[width=\textwidth]{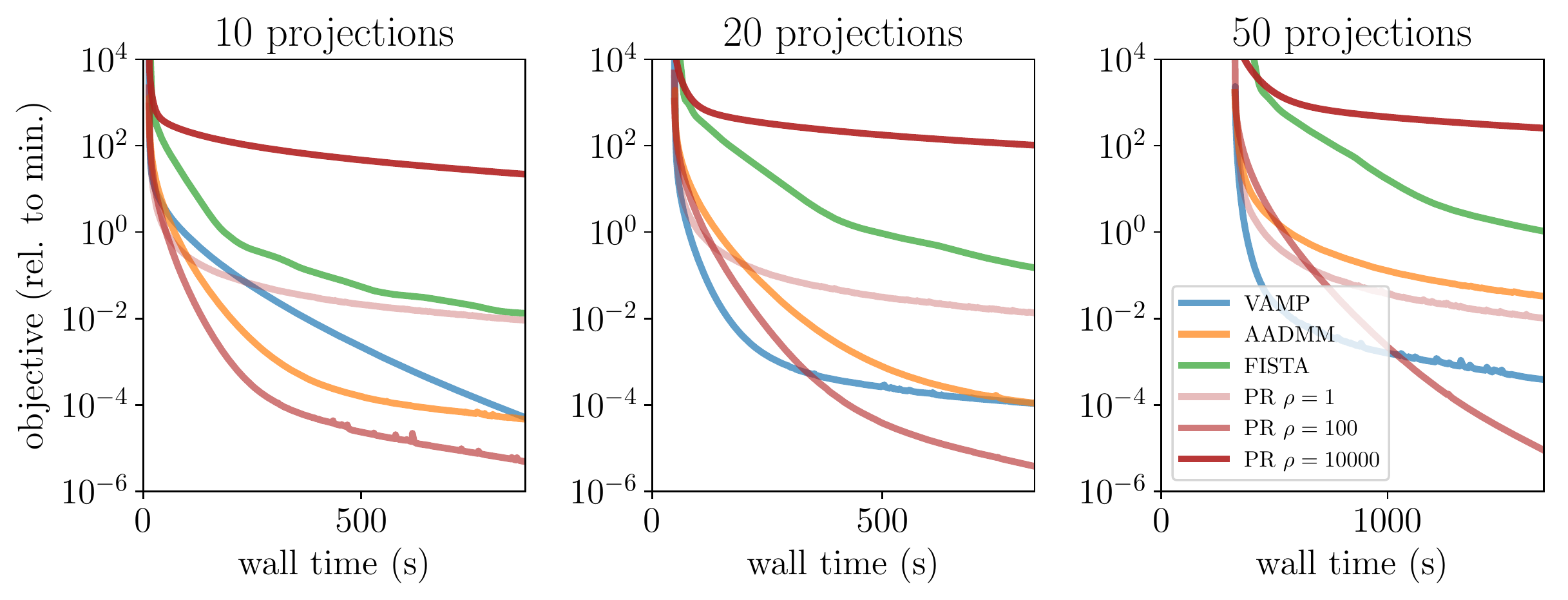}
    \caption{Reconstruction of a Sheep-Logan phantom of size 200x200 ($p =
    40000$) using 10 (left), 20 (center) and 50 (right) equally spaced noisy
    projections ($n = 2000$, $4000$ and $10000$, respectively). A TV penalty
    is used with $\lambda = 1.0$. For 10 and 20 projections, VAMP and AADMM
    are competitive with each other and both faster than FISTA; for 50
    projections, VAMP seems to be considerably faster. Meanwhile, PRS
    performance is highly dependent on the choice of the stepsize $\rho$.
    Note that preprocessing time is not discounted for any of the
    approaches, and that a larger number of projections (i.e.  larger $n$)
    incurs bigger preprocessing times.}
        \label{fig:tomo}
\end{figure}

\subsection{Reconstruction in tomography}

\begin{figure}[ht]
    \centering
    \includegraphics[width=\textwidth]{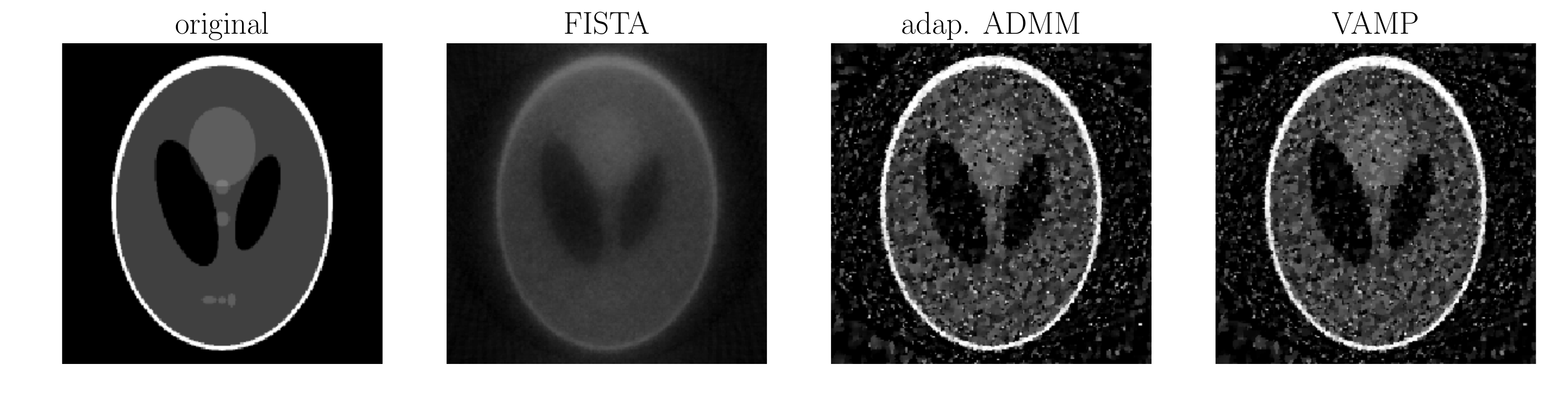}
    \label{fig:phantom}
    \caption{Estimates obtained by the different approaches 10 seconds after
    preprocessing is finished, using 50 noisy projections.}
\end{figure}

A second experiment consists in reconstructing a two-dimensional image from
a number of tomographic projections. We use a Shepp-Logan phantom of dimensions
200x200, so that $p = 40000$, and perform the reconstruction for 10 ($n =
2000$), 20 ($n = 4000$) and 50 ($n = 10000$) equally spaced projections. Moreover, we add
a 1\% SNR noise to the projections. In other words, the tomographic
projections are generated as $\bm{y} = \bm{A} \bm{x} + \mathcal{N} (0,
\sigma^2)$, where $\bm{A}$ is the two-dimensional Radon transform and $\sigma^2 =
0.01 \cdot \| \bm{A} \bm{x} \|_2^2 / n$. In our experiments, we explicitly
generate the Radon matrix, that is, we do not use its operator form.

The regularization parameter is set to $\lambda = 1.0$, and the relaxation
parameter to $\gamma=0.6$.
While for the reconstruction using 20 and 50 projections VAMP is faster than the
remaining algorithms, for 10 projections it is beaten by adaptive-step ADMM,
being, however, still competitive with FISTA. Finally, we show the performance
of Peaceman-Rachford for different values of stepsize $\rho$; while it
might eventually perform better than all the other algorithms, its behaviour
is highly dependant on the choice of stepsize.

%% file: conclusion.tex
\section{Conclusion}

We have proposed a VAMP-like iteration for optimizing functions consisting of
linear losses and a given class of non-separable penalties. As
the usual VAMP iteration, it can be seen as a Peaceman-Rachford splitting
with adaptive stepsizes.
Initial experiments reveal that the resulting algorithm
is competitive with, and faster than other typically employed optimization
schemes, namely ADMM and FISTA. 
While further experiments
should be performed in order to assess the superiority of the algorithm in
more general settings, we believe that the results displayed 
demonstrate its potential for real-world applications.

Our analysis also elucidates the connection between message-passing
algorithms and other known optimization schemes, expanding on the remark
of \cite{fletcher2016expectation}. In particular, the Peaceman-Rachford
splitting is recovered if variances/stepsizes are kept fixed.  Further
exploring this connection should lead to improved inference and optimization
algorithms, and hopefully also to convergence proofs, which are
typically difficult to obtain for message-passing algorithms.

There are other EC/VAMP adaptations able to deal with constraints of the form
$\bm{z} = \bm{K} \bm{x}$, but these are typically used on the loss, not the penalty
\cite{schniter2016vector,he2017generalized}. They could be integrated into our
framework to study losses other than quadratic. An alternative
would be using these adaptations on the penalty instead, by artificially
introducing new effective samples that mimic the penalty as an additional
loss term; this has been explored for AMP in
\cite{borgerding2015generalized,barbier2016scampi}, and is not as
straightforward as the construction we use in this paper.

Yet another perspective is adapting the algorithm for online inference,
using the scheme described in \cite{manoel2017streaming}. Interestingly,
this is similar to what is already done to adapt ADMM to the online
setting \cite{suzuki2013dual}. Finally, we leave for future work comparing
our approach to other recent promising alternatives, in particular
that of \cite{kamilov2017parallel}.

%% file: ackno.tex
\section*{Acknowledgments}
This work was supported by LabEx DigiCosme and the DRF Impulsion program. We
also acknowledge the support from the ERC under the European Unions FP7
Grant Agreement 307087-SPARCS and the European Union’s Horizon 2020 Research
and Innovation Program 714608-SMiLe, as well as by the French Agence
Nationale de la Recherche under grants ANR-17-CE23-0023-01 PAIL and
ANR-17-CE23-0011 Fast-Big.

%% file: appendices.tex
\section{Expectation-consistent approximation}
\label{sec:ec}

The expectation-consistent approximation\cite{opper2005expectation} is defined
for a probability distribution of the form
\begin{equation}
    P(\bm{x}) = \frac{1}{\mathcal{Z}} P_\ell (\bm{x}) \, P_r (\bm{x}).
\end{equation}
It seeks to approximate (minus) the log-partition function $-\log
\mathcal{Z} = -\log \int d\bm{x} \, P_\ell (\bm{x}) \, P_r (\bm{x})$ by
means of the following objective, denoted the \emph{expectation-consistent
free energy}
\begin{equation}
    \mathcal{F} [Q_\ell, Q_r] = -\log \int d\bm{x} \, P_\ell (\bm{x}) \, Q_r (\bm{x})
    - \log \int d\bm{x} \, Q_\ell (\bm{x}) \, P_r (\bm{x})
    + \log \int d\bm{x} \, Q_\ell (\bm{x}) \, Q_r (\bm{x}).
\end{equation}
If there are no restrictions on $Q$, then $\mathcal{F}$ is minimized for
$Q_\ell = P_\ell$, $Q_r = P_r$. What we want however is to have a tractable
approximation, so the $Q$ are typically replaced by distributions in the
exponential family. We use a (unnormalized) Gaussian distribution, written
in canonical form as
\begin{equation}
    Q_{\ell, r} (\bm{x}) = e^{-\frac12 \rho_{\ell, r} \bm{x}^T \bm{x}
        + \bm{u}_{\ell, r}^T \bm{x}}.
\end{equation}
In our case, $P_\ell (\bm{x}) \propto e^{-\frac12 \| \bm{y}
- \bm{A} \bm{x} \|_2^2}$ and $P_r (\bm{x}) = \prod_{j = 1}^p e^{-\lambda f(x_j)}$.
Combined to the explicit form of $Q$ above this gives
\begin{equation}
    \begin{aligned}
        &\mathcal{F} (\bm{u}_\ell, \bm{u}_r, \rho_\ell, \rho_r) = 
        \frac12 \bm{y}^T \bm{y} - \frac12 (\bm{A}^T \bm{y} + \bm{u}_r)^T
        (\bm{A}^T \bm{A} + \rho_r I_p)^{-1} (\bm{A}^T \bm{y} + \bm{u}_r) + \\
        &\kern5em + \frac12 \log \det (\rho_r I_p + \bm{A}^T \bm{A}) -
        \sum_{j = 1}^p \log \int dx_j e^{-\lambda f(x_j)} e^{-\frac12 \rho_\ell x_j^2 +
        u_{\ell k} x_j} + \\
        &\kern5em + \frac12 \sum_{j = 1}^p \left[ \frac{u_{rk} + u_{\ell k}}{\rho_r
        + \rho_\ell} - \log (\rho_r + \rho_\ell) \right].
    \end{aligned}
\end{equation}
There are different ways of looking for local minima of $\mathcal{F}$
\cite{opper2005expectation}; the vector AMP iteration \cite{rangan2017vector}
is one of them. It consists in the following fixed-point iteration

\begin{equation}
    \begin{aligned}
        &\bm{u}_{\ell}^{t + 1} = \frac{\mathbb{E}_\ell^t
        \bm{x}}{\langle \Var_\ell^t (\bm{x}) \rangle} -
        \bm{u}_{r}^t, \qquad
        \rho_{\ell}^{t + 1} = \frac{1}{\langle \Var_\ell^t
        (\bm{x}) \rangle} - \rho_r^t, \\
        &\bm{u}_r^{t + 1} = \frac{\mathbb{E}_r^t
        \bm{x}}{\langle\Var_r (\bm{x}) \rangle} -
        \bm{u}_\ell^t, \qquad
        \rho_r^{t + 1} = \frac{1}{\langle \Var_r^t
        (\bm{x}) \rangle} - \rho_\ell^t,
    \end{aligned}
\end{equation}
where we denote by $\mathbb{E}_{\ell, r}^t$ the expectation w.r.t.
the tilted distributions $\tilde{Q}_{\ell, r}^t (\bm{x}) \propto P_{\ell, r}
(\bm{x}) Q_{\ell, r}^t (\bm{x})$, and by $\Var_{r, \ell}^t (\bm{x})$ the variance
of these distributions. In particular
\begin{equation}
    \mathbb{E}_{\ell}^t \bm{x} = (\bm{A}^T \bm{A} + \rho_r^t I_p)^{-1} (\bm{A}^T \bm{y} + \bm{u}_r^t),
    \qquad
    \langle \Var_\ell^t (\bm{x}) \rangle = \frac{1}{p} \Tr (\bm{A}^T \bm{A} + \rho_r^t I_p)^{-1},
\end{equation}
and, by defining $z (u, \rho) = \int dx e^{-\lambda f(x)} e^{-\frac12 \rho
x^2 + u x}$,
\begin{equation}
    \mathbb{E}_r^t x_j = \frac{\partial}{\partial u} \log z (u, \rho)
        \Big|_{u_{\ell k}^t, \rho_\ell^t},
    \qquad
    \langle \Var_r^t (\bm{x}) \rangle = \frac{1}{p} \sum_{j = 1}^p
        \frac{\partial^2}{\partial u^2} \log z (u, \rho) \Big|_{u_{\ell k}^t,
        \rho_\ell^t},
    \label{eq:ec_avgr}
\end{equation}

Note that upon convergence one achieves the stationary conditions $\nabla
\mathcal{F} = 0$
\begin{equation}
    \bm{u}_\ell + \bm{u}_r = \frac{\mathbb{E}_\ell \bm{x}}{\langle
    \Var_\ell (\bm{x}) \rangle} = \frac{\mathbb{E}_r
    \bm{x}}{\langle \Var_r (\bm{x}) \rangle}, \qquad
    \rho_\ell + \rho_r = \frac{1}{\langle \Var_\ell (\bm{x}) \rangle} =
    \frac{1}{\langle \Var_r (\bm{x}) \rangle}.
\end{equation}
In other words, the first two moments of $\tilde{Q}_\ell$ and $\tilde{Q}_r$ must be
consistent with those of the approximating posterior $Q (\bm{x}) \propto
Q_\ell (\bm{x}) Q_r (\bm{x})$.

\subsection{MAP estimate}

In order to go from the equations above to (\ref{eq:vamp1})-(\ref{eq:vamp3}),
we perform two additional steps. First, note that one does not need to work
with both $\ell$ and $r$ variables; by denoting $\bm{u} \equiv \bm{u}_r$,
we rewrite the iteration as
\begin{equation}
    \bm{u}^{t + 1} = \bm{u}^t + \left( \frac{\mathbb{E}_r^t
        \bm{x}}{\langle \Var_r^t (\bm{x}) \rangle} -
        \frac{\mathbb{E}_\ell^t \bm{x}}{\langle \Var_\ell^t (\bm{x})
        \rangle} \right),
\end{equation}
and analogously for $\rho$, thus obtaining (\ref{eq:vamp2}). Moreover, since
we are interested in the MAP estimate, we use the following trick: we
consider instead a distribution $P (\bm{x}) \propto [ P_\ell (\bm{x}) \, P_r
(\bm{x}) ]^\beta$ and take the limit $\beta \to \infty$, thus making the
distribution concentrate around its mode; we then look at $\frac{1}{\beta}
\log \mathcal{Z}$. By rescaling $\bm{u}_{\ell, r} \to \beta \bm{u}_{\ell, r}$
and $\rho_{\ell, r} \to \beta \rho_{\ell, r}$, we ensure that most equations
remain the same. The only difference is in the second integral in
$\mathcal{F}$, which can be performed using the Laplace method to yield
\begin{equation}
    \frac{1}{\beta} \log z(u, \rho) =
    \frac{1}{\beta} \log \int dx_j e^{-\beta \left\{ \lambda f(x) +
        \frac12 \rho x - u x \right\}} \xrightarrow{\beta \to \infty}
     \frac{u^2}{2 \rho} - \rho \mathcal{M}_{\lambda f /
         \rho_\ell} (u / \rho),
\end{equation}
where $\mathcal{M}_f (v) = \min_{x} \Big\{ f(x) + \frac12 (x - v)^2 \Big\}$ is
the Moreau envelope of $f$. Given that $\frac{\partial}{\partial v} \mathcal{M}_f (v) 
= x - \prox_f (v)$, we get using (\ref{eq:ec_avgr}) that
\begin{equation}
    \mathbb{E}_r^t \bm{x} = \prox_{\lambda f / \rho_\ell^t} (\bm{u}_{\ell} /
    \rho_\ell) \equiv \eta_{\lambda f / \rho_\ell^t} (\bm{u}_\ell / \rho_\ell),
\end{equation}
and also that $\langle \Var_r^t (\bm{x}) \rangle = \rho_\ell^{-1}
\langle \nabla \eta_{\lambda f / \rho_\ell^t} (\bm{u}_\ell / \rho_\ell)
\rangle$, leading to (\ref{eq:vamp1}) and (\ref{eq:vamp3}).

\section{Adapting the EC approximation to TV penalties}
\label{sec:adapt}

In order to adapt iteration (\ref{eq:vamp1})-(\ref{eq:vamp2}) to total variation
penalties, we look at the expectation-consistent (EC) approximation
\cite{opper2005expectation} in closer detail. We first define the following
distribution
\begin{equation}
    P (\bm{x}) \propto e^{-\frac12 \| \bm{y} - \bm{A} \bm{x} \|_2^2} \,
        \prod_{k = 1}^r e^{-\lambda f\big( (\bm{K} \bm{x})_k \big)}.
\end{equation}
Problem (\ref{eq:prob_nonsep}) is then recovered when considering the MAP
estimator for $\bm{x}$, given by the mode of $P (\bm{x})$. Next we introduce
a variable $\bm{z} = \bm{K} \bm{x}$, and rewrite the distribution above as
\begin{equation}
    P (\bm{x}, \bm{z}) \propto e^{-\frac12 \| \bm{y} - \bm{A} \bm{x} \|_2^2} \,
        \delta(\bm{z} - \bm{K} \bm{x}) \prod_{k = 1}^r e^{-\lambda f\big( z_k \big)}.
\end{equation}
so that the marginal distribution of $\bm{z}$ becomes
\begin{equation}
    P (\bm{z}) = \frac{1}{\mathcal{Z}} \, \bigg[ \! \int d\bm{x} \, e^{-\frac{1}{2}
    \| \bm{y} - \bm{A} \bm{x} \|_2^2} \, \delta(\bm{z} - \bm{K} \bm{x}) \bigg] \, \prod_{k = 1}^r
    e^{-\lambda f(z_k)}.
    \label{eq:p_z}
\end{equation}
We apply the EC scheme described above for $P (\bm{z})$, by setting
\begin{equation}
    P_\ell (\bm{z}) \propto \int d\bm{x} \, e^{-\frac12 \| \bm{y} - \bm{A} \bm{x} \|_2^2}
        \, \delta (\bm{z} - \bm{K} \bm{x}), \qquad
    P_r (\bm{z}) \propto \prod_{k = 1}^r e^{-\lambda f(z_k)},
\end{equation}
thus yielding the following expectation-consistent free energy
\begin{equation}
    \mathcal{F} [Q_\ell, Q_r] = -\log \int d\bm{x} \, e^{-\frac12 \| \bm{y} - \bm{A} \bm{x} \|_2^2}
    \, Q_r (\bm{K} \bm{x}) - \log \int d\bm{z} \, Q_\ell (\bm{z}) P_r (\bm{z})
    + \log \int d\bm{z} \, Q_\ell (\bm{z}) Q_r (\bm{z}),
\end{equation}
We once again parametrize these distributions by canonical-form Gaussians,
$Q_{\ell, r} (\bm{z}; \bm{u}_{\ell, r}, \rho_{\ell, r}) =
e^{-\frac{\rho_{\ell, r}}{2} \bm{z}^T \bm{z} + \bm{u}_{\ell, r}^T \bm{z}}$,
leading to the fixed-point iteration (\ref{eq:tvamp1})-(\ref{eq:tvamp3}).

\section{VAMP iteration for $n \ll p$}
\label{sec:app_nllp}

In the $n \ll p$ setting, iteration (\ref{eq:vamp1}) can be made
efficient by combining the Woodbury formula with the eigendecomposition of
$\bm{A} \bm{A}^T = \bm{U} \operatorname{diag} (\bm{d}) \bm{U}^T$. We first use the Woodbury
formula to replace the iteration on $\bm{x}$ by
\begin{equation}
    \bm{x}^{t} = \frac{\bm{u}^t}{\rho^t} + \bm{A}^T (\bm{A} \bm{A}^T + \rho^t I_n)^{-1}
    \bigg( y - \bm{A} \frac{\bm{u}^t}{\rho^t} \bigg),
\end{equation}
and then insert the eigendecomposition
\begin{equation}
    \bm{x}^{t} = \frac{\bm{u}^t}{\rho^t} + \bm{A}^T \bm{U} \bigg\{ \bigg[ \bm{U}^T
    \bigg( y - \bm{A} \frac{\bm{u}^t}{\rho^t} \bigg) \bigg] \oslash
    (\bm{d} + \rho^t) \bigg\},
    \label{eq:xupd_woodbury}
\end{equation}
where we have used the symbol $\oslash$ to denote the Hadamard or elementwise
division between two vectors. Note that evaluating the expression above
consists only in performing matrix-vector products and other vector-vector
operations, and can thus be performed in $O(np)$.

Finally, iteration (\ref{eq:vamp3}) can be replaced by
\begin{equation}
    \sigma_x^t = \frac{1}{n} \sum_{i = 1}^n \frac{1}{d_i + \rho^t}.
\end{equation}

\subsection{Further simplifications for TV penalty}
\label{sec:simp_tv}

For the non-separable penalty (\ref{eq:prob_nonsep}), the same trick can be
applied to yield
\begin{equation}
    \bm{x}^t = (\bm{K}^T \bm{K})^{-1} \frac{\bm{u}^t}{\rho^t} + (\bm{K}^T \bm{K})^{-1} \bm{A}^T \bm{U}
    \bigg\{ \bigg[ \bm{U}^T (\bm{y} - \bm{A} (\bm{K}^T \bm{K})^{-1} \frac{\bm{u}^t}{\rho^t} \bigg]
    \oslash (\bm{d} + \rho^t) \bigg\},
\end{equation}
and, by further using the eigendecomposition of $\bm{K}^T \bm{K} = \bm{F}^T \bm{\Lambda} \bm{F}$
\begin{equation}
    \bm{x}^t = \bm{F}^T \bm{\Lambda}^{-1} \Bigg\{ \bm{F} \frac{\bm{u}^t}{\rho^t} + \tilde{\bm{A}}^T \bm{U}
    \bigg\{ \bigg[ \bm{U}^T (\bm{y} - \tilde{\bm{A}} \bm{\Lambda}^{-1} \bm{F} \frac{\bm{u}^t}{\rho^t} \bigg] \oslash
    (\bm{d} + \rho^t) \bigg\} \Bigg\},
\end{equation}
for $\tilde{\bm{A}}^T = \bm{F} \bm{A}^T$. In particular for the TV penalty, $\bm{K} = \nabla$ is the gradient
operator and $\bm{K}^T \bm{K} = \Delta$ is the Laplacian, which being a circulant matrix
has $\bm{F}$ given by the Fourier transform. We can then replace $\bm{F}$ by the FFT,
so that the expression above can be evaluated almost as efficiently as
(\ref{eq:xupd_woodbury}).

The iteration on $\sigma_x$ is written in this case as
\begin{equation}
    \sigma_x^t = \frac{1}{\rho^t} \left[\frac{1}{d} - \frac{1}{nd} \sum_{i
    = 1}^{nd} \frac{d_i}{d_i + \rho^t} \right] = \frac{1}{nd} \sum_{i =
    1}^{nd} \frac{1}{d_i + \rho^t} - \frac{d - 1}{d} \frac{1}{\rho^t},
\end{equation}
where $d$ is the dimension of the TV operator, i.e. $\bm{K} = \nabla$ is a $n d \times n$ matrix.